\documentclass[letterpaper, 10 pt, conference]{ieeeconf} 
\IEEEoverridecommandlockouts 
\usepackage{graphicx}
\usepackage{amssymb}
\usepackage{amsmath}
\usepackage{threeparttable}
\usepackage{algorithm}
\usepackage{algorithmic}
\usepackage{cite}
\usepackage{url}
\usepackage{color}
\usepackage{bm}
\usepackage{footnote}

\makeatletter
\let\NAT@parse\undefined

\newcommand{\Rmnum}[1]{\expandafter\@slowromancap\romannumeral #1@}
\makeatother

\usepackage[colorlinks, linkcolor=black, anchorcolor=black, citecolor=black]{hyperref}

\title
{\LARGE \bf
CMTS: Conditional Multiple Trajectory Synthesizer \\
for Generating Safety-critical Driving Scenarios
}

\author{Wenhao Ding$^{1}$, Mengdi Xu$^{1}$ and Ding Zhao$^{1}$% <-this % stops a space
\thanks{$^{1}$Wenhao Ding, Mengdi Xu and Ding Zhao are with the Department of Mechanical Engineering, Carnegie Mellon University, Pittsburgh, PA 15213, USA. {\tt\small wenhaod@andrew.cmu.edu}, {\tt\small mengdixu@andrew.cmu.edu}, {\tt\small dingzhao@cmu.edu}}%
}

\begin{document}
\maketitle

%%%%%%%%%%%%%%
\begin{abstract}

Naturalistic driving trajectories are crucial for the performance of autonomous driving algorithms. However, most of the data is collected in safe scenarios leading to the duplication of trajectories which are easy to be handled by currently developed algorithms. When considering safety, testing algorithms in near-miss scenarios that rarely show up in off-the-shelf datasets is a vital part of the evaluation.
As a remedy, we propose a near-miss data synthesizing framework based on Variational Bayesian methods and term it as Conditional Multiple Trajectory Synthesizer (CMTS). We leverage a generative model conditioned on road maps to bridge safe and collision driving data by representing their distribution in the latent space. By sampling from the near-miss distribution, we can synthesize safety-critical data crucial for understanding traffic scenarios but not shown in neither the original dataset nor the collision dataset.
Our experimental results demonstrate that the augmented dataset covers more kinds of driving scenarios, especially the near-miss ones, which help improve the trajectory prediction accuracy and the capability of dealing with risky driving scenarios.

\end{abstract}
%%%%%%%%%%%%%%

%%%%%%%%%%%%%%
\section{Introduction}
%%%%%%%%%%%%%%

\begin{figure}[t]
\centering
\includegraphics[width=8.5cm]{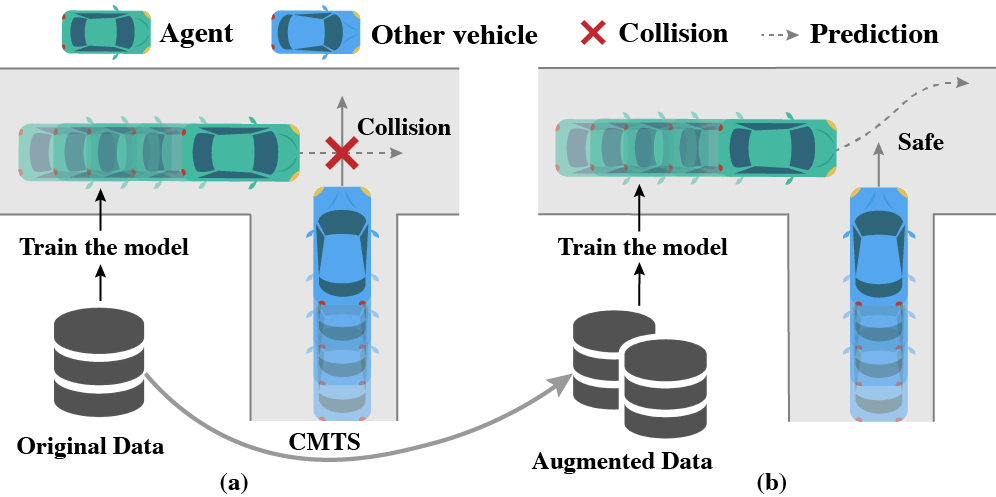}
\caption{In a safety-critical situation, one trajectory prediction algorithm outputs a collision prediction when trained with the original dataset as shown in (a), but a safe prediction when trained with the dataset augmented by CMTS as in (b).}
\label{overview}
\end{figure}

Data acquisition vehicles are running on roads and different autonomous driving research institutes have already released their datasets containing millions of data \cite{56}\cite{64}. However, most of the time vehicles drive safely without threats in the real world and the collected data is repetitive for belonging to the same scenario. 
%For the autonomous driving algorithms, the richer the types of scenes in the training dataset, the more driving scenarios, especially near-miss scenarios, the algorithm can handle (Fig.~\ref{overview}). %
With a small number of near-miss scenarios, algorithms tend to overfit on safe scenarios and are thus hard to generalize to risky ones as shown in Fig.~\ref{overview}. Unfortunately, few released datasets contain a fair fraction of near-miss data. In addition to collecting risk data by driving for a longer time, another more efficient way is directly generating the data we need.

Recently, generative models have been widely used in the image field, including Generative Adversarial Networks (GAN)\cite{44}, Variational Auto-encoder (VAE)\cite{6}, Glow\cite{50}, etc. The essence of these models is first fitting a distribution based on collected data, and then generating new data by sampling from this distribution. To successfully get the parameters shaping the distribution of a dataset, a large amount of data is required. In our case, the released datasets provide us adequate data to use the generative model to augment the original dataset.

\begin{figure*}[t]
\centering
\includegraphics[width=17cm]{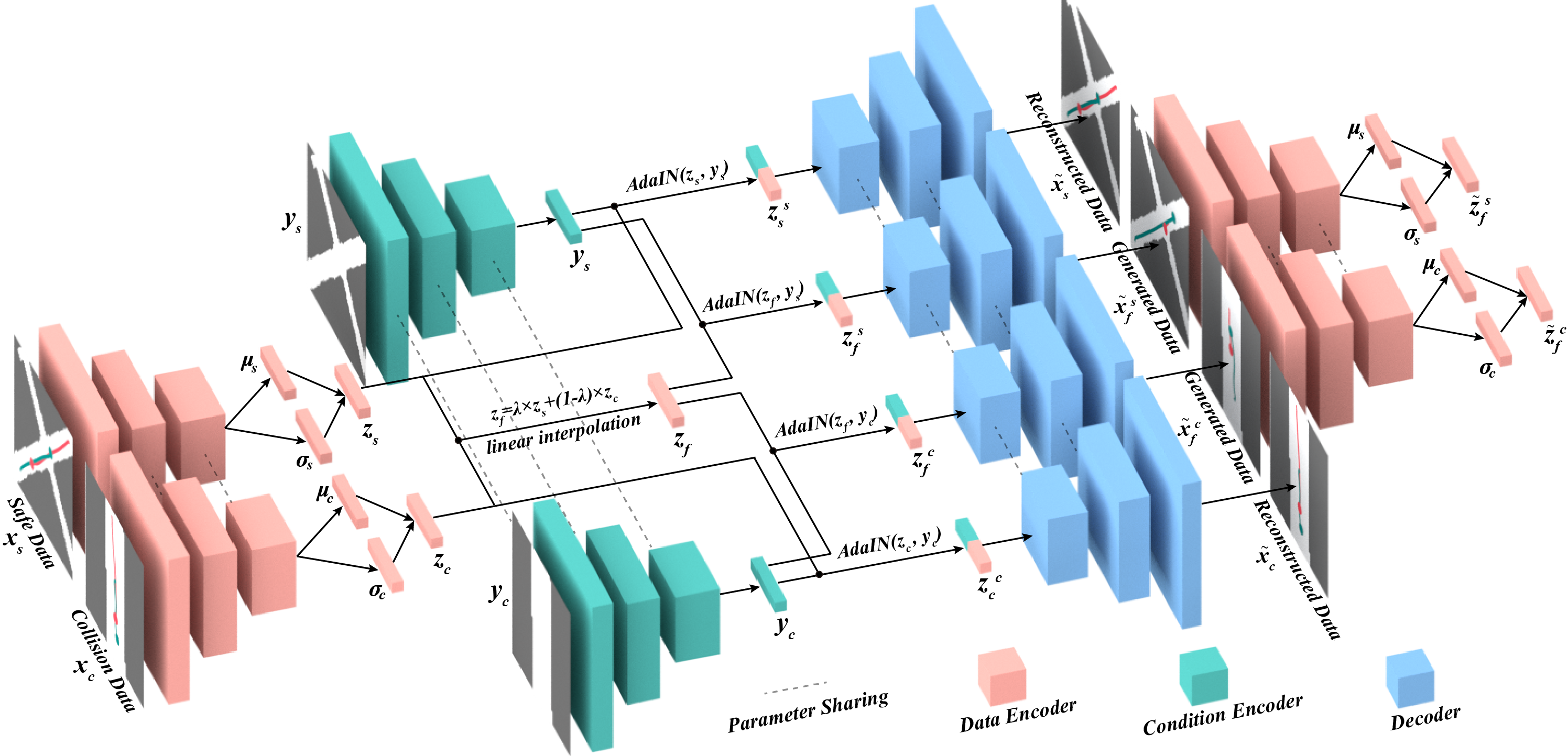}
\caption{Structure of CMTS which consists of three modules: a sequential trajectory encoder (red), a conditional map encoder (green), and a decoder that generates trajectory (blue). Dimensions of layers can be found in the supplementary material.}
\label{structure}
\end{figure*}

Now that we are able to utilize the generative models, how do we generate near-miss data? First of all, manifold learning shows that high-dimensional data (images, point clouds, trajectories, etc.) can be expressed in a low-dimensional space, which makes it possible to manipulate data in a more semantic way. Secondly, the data collected in the real world and via simulators can easily provide safe and collision driving data. 

Based on the aforementioned two preconditions, we propose a variational Bayesian framework named CMTS as shown in Fig.~\ref{structure} to synthesis near-miss driving data from naturalistic safe trajectory data and artificial collision trajectory data. The framework mainly contains 3 steps which are (a) encoding the distributions of safe data and collision data into a low-dimensional latent space, (b) embedding the road information in this latent space, (c) synthesising risk scenarios from the interpolated intermediate distribution. Before getting the representation in low-dimensional space, we use a style transfer method to separate the road information from the driving behavior information.
%, and then project the driving behavior information into a low-dimensional latent space. 
%The prior distribution of driving behavior representation in the latent space is set as a multivariate normal distribution, which makes it easy to interpolate when given the distributions of two datasets.

To evaluate the effectiveness of our framework, we leverage statistical indicators to measure the complexity of the augmented dataset and the original dataset. The experimental result demonstrates that our augmented data is richer than the previous data. Moreover, the augmented dataset improves the accuracy of trajectory prediction algorithms.

The contribution of this paper is three-fold:
\begin{itemize}
	\item A variational Bayesian framework with latent space interpolation to fuse information of two domains;
	\item Rare near-miss driving data generated by leveraging the proposed framework using public driving datasets.
	\item The accuracy improvement of trajectory prediction algorithms and the capability enhancement to deal with risky scenarios when trained with the augmented dataset. 
\end{itemize}

%%%%%%%%%%%%%%%
\section{Related Work}
%%%%%%%%%%%%%%%

\subsection{Trajectory and Behavior Prediction}

To safely navigate on roads, being able to predict the dynamics of the driving environment is crucial. Many algorithms have already been proposed to predict the motion of pedestrians \cite{16, 17, 18, 19} and vehicles \cite{20, 21, 22, 24}. Considering the correlation between the trajectories of multiple agents, \cite{16} proposed Social-LSTM to achieve information communication by sharing agents' hidden states. \cite{20} proposed a convolutional social layer to encode interactive information. Instead of predicting deterministic results as in \cite{16} and \cite{20}, \cite{18} output a distribution which describes the uncertainty by its variance. In addition, \cite{23, 24, 27, 63} provided the multi-modal predictions with probabilities for each modality. \cite{22} proposed a hierarchical structure, using high-level intention predictions to guide the prediction of the trajectory.

Although modifying the structure of forecasting algorithms helps improve the accuracy, the lack of complex vehicle interaction data, as claimed in \cite{21}, is an even more critical factor limiting the prediction algorithm. As far as we know, only a few works focused on the data augmentation or the complex trajectory generation of near-miss scenarios. \cite{61} proposed to add perturbations to increase data complexity, but did not take the encounter scenarios into account. \cite{62} leveraged VAE to generate new trajectories from the original dataset. However, neither \cite{61} nor \cite{62} considered the road constraints during synthesizing.
%and this paper is the first try to use generative models to study this problem.

\subsection{Deep Generative Models}

In recent years, GAN \cite{44} shows its privilege in image generation which stimulates the research trend of using deep generative models. In general, the current popular deep generative models fall into three categories: (a) flow-based models that directly calculate the likelihood \cite{48, 49, 50}; (b) VAEs that calculate the approximate likelihood using variational inference \cite{6, 46, 47}; (c) GANs \cite{44, 45} that implicitly calculate the likelihood with adversarial training. Despite the different ways of optimization, all these three types of models use existing samples to fit the distribution of the data. After getting the distribution, the most common application is to draw samples from it to generate new data, such as painting \cite{52} and music \cite{53}. Because of the powerful data generation capability, these models are extensively used in data augmentation \cite{54, 55}.

Although GAN is good at image generation, the instability of adversarial training, which is heavily influenced by hyper-parameters and network structures, has not been completely solved so far. The computational cost in the flow-based model also limits its widespread use. In contrast, VAE is relatively more stable and efficient. Although VAE may have a certain bias for the estimation of the likelihood, its Gaussian prior assumption for the latent variable is useful for controlling the properties of the generated data, especially for the latent space interpolation.

\subsection{Interpolation of Generative Models}

Latent space interpolation is an advantageous method to modify data samples. A typical application is editing face semantic attributes (gender, smile, glass, etc) \cite{13, 31, 32, 35, 36, 37, 38}. Assuming that only one attribute is different between two sets of samples, the high-dimensional data samples are first projected into the low-dimensional hidden space. After linear interpolation, the features are then projected back to initial data space. In this process, the occurrence of one specific attribute is controlled by adjusting the interpolation coefficient. This kind of method usually requires interpolation inside the manifold, therefore how to avoid interpolation results outside the domain becomes a key point. Some researchers proposed to use Riemannian metric instead of linear interpolation in \cite{41}, while others changed the shape of manifolds using adversarial training to satisfy the nature of convex sets \cite{29, 34}. Changing the shape of the latent space by adding constraints in the training process also helps improve the performance of semi-supervised learning methods\cite{42, 43}.

Although interpolation is a favorable tool, it has a strong assumption that there is only one attribute under control. If the data contains two or more attributes, the result of the interpolation is unpredictable. For example, the real-world driving trajectory contains the road information. Interpolating two trajectories directly in the latent space will also blend different road constraints. Thence we regard the road map as the style information and separate it from others in the trajectory via the style transfer method.

\subsection{Style Transfer}

The most intuitive examples of style transfer are in the computer vision area. The requirements of preserving the objects and the painting layouts while changing the style are fulfilled using style transfer \cite{7, 8, 9, 10, 11, 14}. Essentially, a style can be viewed as a condition where a painting is generated. The dominant method to achieve style transfer is the AdaIN proposed in \cite{3}, which first uses the mean and variance to model the condition, and then manipulates these two statistics to change the style \cite{12}. In this work, we follow this convention and also regard road conditions as styles.

%%%%%%%%%%
\section{Method}
%%%%%%%%%%

In this section, we describe the framework structure of CMTS by providing the VAE preliminary, the framework overview and the detailed explanation of core modules.

\subsection{Preliminary} \label{sec_A}

The VAE \cite{6}\cite{5} is a directed graphical model that contains two parts, inference process and generative process. In generative process, the latent variable $z$ is generated from the prior distribution $p_{\theta}(z)$ characterized by $\theta$ and the data $x$ is generated by $p_{\theta}(x|z)$. The parameter of the generative part $\theta$ is obtained through optimization. One direct way to optimize the auto-encoder is to maximize the likelihood $log\,p_{\theta}(x)$ of all data points.
\begin{equation}
\begin{aligned}
log\,p_{\theta}(x)&=E_{q_{\phi}(z|x)}\left[log\,p_{\theta}(x)\right] \\
                  &=E_{q_{\phi}(z|x)}\left[log\frac{p_{\theta}(x, z)\,q_{\phi}(z|x)}{q_{\phi}(z|x)\,p_{\theta}(z|x)}\right] \\
                  &=\underbrace{E_{q_{\phi}}\left[log\frac{p_{\theta}(x, z)}{q_{\phi}(z|x)}\right]}_{=\mathcal{L}(x)} + 
                  	\underbrace{KL\left(q_{\phi}(z|x)||p_{\theta}(z|x)\right)}_{\geq0},
\end{aligned}
\label{eq1}
\end{equation}
where KL is the Kullback-Leibler divergence (KLD), and $q_{\phi}(z|x)$ is the variational approximation of $p_{\theta}(z|x)$. Since the KL term is non-negative, $\mathcal{L}_{\theta,\phi(x)}$ is the lower bound of the likelihood $log\,p_{\theta}(x)$, which is also called the evidence lower bound (ELBO).
\begin{equation}
\begin{aligned}
\mathcal{L}(x)&=E_{q_{\phi}}\left[-log\,q_{\phi}(z|x) + log\,p_{\theta}(x,z)\right] \\
                            &=-KL(q_{\phi}(z|x)||p_{\theta}(z)) + E_{q_{\phi}}\left[log\,p_{\theta}(x|z)\right] \\
\end{aligned}
\label{eq2}
\end{equation}

Therefore, our goal to maximize the likelihood is equivalent to maximize the ELBO in (\ref{eq2}), the second term of which is denoted as the negative reconstruction error in the terminology of auto-encoder.

\subsection{Framework Overview} \label{sec_B}

The proposed CMTS is displayed in Fig.~\ref{structure}. Our framework contains three parts: (a) a Gated Recurrent Unit (GRU) data encoder that encodes two sequence data (one from original dataset, another from collision dataset) into a latent space; (b) a convolutional condition encoder that encodes the road map into a multivariate Gaussian distribution; (c) a GRU decoder that re-projects the combination of sequence and road map back to the high-dimensional data space. How to combine the sequence and map information is detailed in Sec.~\ref{sec_C} and the random interpolation process during the training process is described in Sec.~\ref{sec_D}.

The overall loss function is:  
\begin{equation}
\mathcal{L}_{CMTS}(x|y)=\alpha (\mathcal{L}_r^s+\mathcal{L}_r^c) + \beta (\mathcal{L}_{KL}^s+\mathcal{L}_{KL}^c) + \gamma \mathcal{L}_f,
\label{eq3}
\end{equation}
where $y$ is the condition or style represented by grid map images and $\alpha$, $\beta$ and $\gamma$ are weights of different parts of the loss. $\mathcal{L}_r^s$ and $\mathcal{L}_r^c$ are the reconstruction errors of the safe and the collision dataset, respectively. $\mathcal{L}_{KL}^s$ and $\mathcal{L}_{KL}^c$ are the KLD of the safe and the collision dataset, respectively. ${L}_f$ is the consistent regularization introduced in Sec.~\ref{sec_E}, which helps improve the performance of interpolation.

\subsection{Merging Conditions with Style Transfer} \label{sec_C}

The derivation of conditional VAE \cite{1} is similar to (1). The only difference is that both the generative and inference model are conditioned by $y$. In CMTS, the condition $y$ denotes the information of the road constraint, which is already contained in the corresponding data point $x$, thus we assume that $q_{\phi}(z|x,y)=q_{\phi}(z|x)$. In addition, we further relax the constraints so that the prior distribution of latent code $z$ is statistically independent of input variables \cite{2}. Therefore, we obtain the optimization function of conditional VAE as
\begin{equation}
\begin{aligned}
\mathcal{L}(x|y)=&-KL(q_{\phi}(z|x, y)||p_{\theta}(z|y)) + E_{q_{\phi}}\left[log\,p_{\theta}(x|z,y)\right] \\
                =&\underbrace{-KL(q_{\phi}(z|x)||p_{\theta}(z))}_{=\mathcal{L}_{KL}}+\underbrace{E_{q_{\phi}}\left[log\,p_{\theta}(x|z,y)\right]}_{=\mathcal{L}_r}.
\end{aligned}
\label{eq4}
\end{equation}

Here, we retain the condition $y$ in $p_{\theta}(x|z,y)$ because we regard the condition as the style of the data which is changed in the generative part. To achieve the style transfer operation, we resort to a prevalent method called AdaIN \cite{3}. Since we suppose the style information has been included in $x$ and $z$, it is reasonable to first remove the original before assigning a new style. The formula of AdaIN is described in (\ref{eq5}).
% \begin{equation}
% \bm{z}_y = AdaIN(\bm{z}_x, \bm{y}) = \sigma(\bm{y})  \frac{\bm{z}_x-\mu(\bm{x})}{\sigma(\bm{x})}+\mu(\bm{y})
% \label{eq5}
% \end{equation}
\begin{equation}
z_y = AdaIN(z_x, y) = {\sigma}(y)  \frac{z_x-{\mu}(x)}{{\sigma}(x)}+{\mu}(y)
\label{eq5}
\end{equation}
where $z_x$ is conditioned by $x$ and $z_y$ is conditioned by $y$. Since the condition in our case is the binary grid map, we use convolution layers to extract features and directly output the ${\mu}(y)$ and ${\sigma}(y)$ of condition $y$.

\subsection{Interpolation during Training} \label{sec_D}

We propose to use linear interpolation during the training stage to obtain the composed feature. We choose VAE as our basic model because its prior distribution of the latent code $z$ is a multivariate Gaussian distribution, which is naturally disentangled. Suppose the latent code of the safe data and the collision data are $z_s$ and $z_c$ respectively, the linear interpolation in latent space will be:
\begin{equation}
z_f = \lambda z_s + (1-\lambda)z_c,
\end{equation}
where $\lambda \in [0,1]$ is the weight controlling the proportion of components. The interpolation distribution is presented as:
\begin{equation}
	p(z_f | \lambda) \sim \mathcal{N}(\lambda \mu_{s}+(1-\lambda)\mu_{c}, {\lambda}^2 \sigma_{s}^2+{(1-\lambda)}^2 \sigma_{c}^2).
\end{equation}
$\lambda$ is drawn from a uniform distribution $U[0, 1]$ for each pair of data points during the training process. In the generating stage, $\lambda$ is fixed and $z_f$ is sent to the decoder to generate near-miss data.

\subsection{Reconstruction of Fusion Samples} \label{sec_E}

The remaining problem is how to measure the reconstruction error of composed latent code $z_f$. For $z_s$ and $z_c$, we directly calculate the difference of $x$ and $q_{\phi}(x|z)$; however, there is no reference for $z_f$ in the dataset. To solve this problem, we propose a consistent regularization from the mutual information view that is similar to \cite{4}. We ensure the reconstruction of $z_f$ by building a unique mapping from $x_f$ to $z_f$ with the encoder $p_{\theta}(x|z)$. From the point of mutual information, there is a unique mapping from $x$ to $z$ if and only if the entropy $H(z|x)=0$. However, it is intractable to access all data pairs to calculate the entropy, thus we use variational inference to obtain the upper bound of $H(z|x)$:
\begin{equation}
\begin{aligned}
H(z|x)&\triangleq-\sum_{z}\sum_{x}p(x)p(z|x)log\,p(z|x) \\
      &\propto-\sum_{z}p(z|x)log\,p(z|x) \\
      %&=-\sum_{z}p(z|x)log\,q(z|x)-\sum_{z}p(z|x)\left[log\frac{p(z|x)}{q(z|x)}\right] \\
      &=H_{p(z|x)}[log\,q(z|x)]-\underbrace{KL(p(z|x)||q(z|x))}_{\geq0} \\
      &\geq H_{\tilde{z}\sim q_{\phi}(z|x), \tilde{x}\sim p_{\theta}(x|\tilde{z})}[log\,q_{\phi}(z|\tilde{x})] \triangleq \mathcal{L}_{f}(z, \tilde{z})
\end{aligned}
\label{eq8}
\end{equation}
In the second step in (\ref{eq8}), we assume $p(x)$ is a uniform distribution, which is reasonable for a dataset without any prior knowledge. The result of (\ref{eq8}) uses the similarity of $z$ and $\tilde{x}$ to represent the entropy $H(z|x)$, providing a consistent regularization term to guarantee the reconstruction of $z_f$. In implementation, we use 2-norm to calculate this $\mathcal{L}_{f}(\cdot, \cdot)$.

\begin{figure}[t]
\centering
\includegraphics[width=8.5cm]{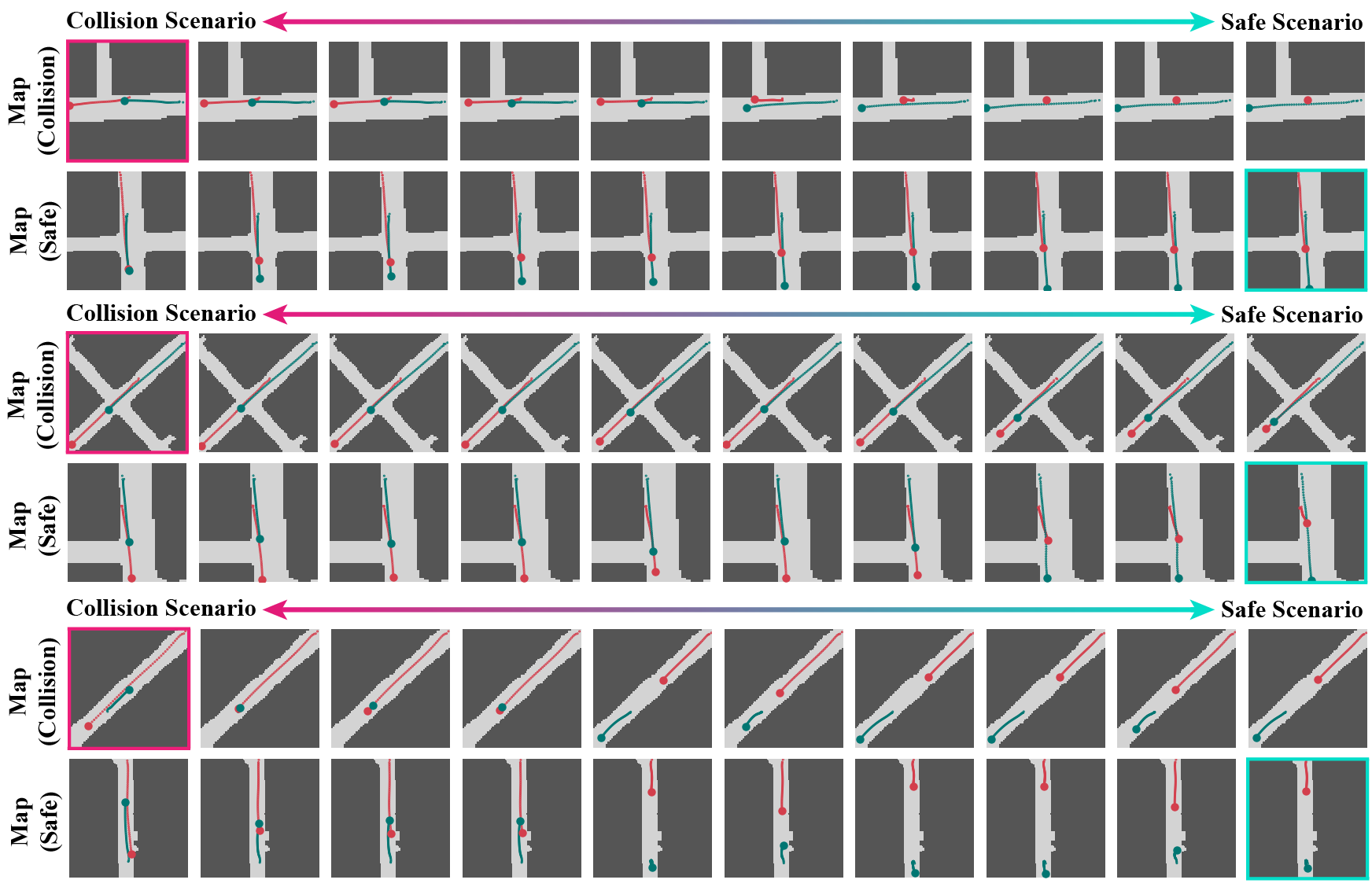}
\caption{Three interpolation examples between safe trajectories and collision trajectories using Argoverse dataset. The images in the pink box (upper left) and the cyan box (lower right) are sampled from the original dataset and the collision dataset, respectively. For each group, the first row shows the results under the map of the safe sample, and the second row shows the results under the map of the collision sample.}
\label{transfer}
\end{figure}

%%%%%%%%%%%%%
\section{Experiment Settings}
%%%%%%%%%%%%%

\subsection{Dataset and Baseline}

To demonstrate the performance of modules in CMTS, we experimented on the following datasets and compared with other data augment algorithms as baselines.

\subsubsection{Lines Dataset} Lines dataset was proposed in \cite{28}. This artificial dataset is simple so that the interpolating smoothness could be easily observed by some metrics such as the \textit{Smoothness} and \textit{Mean Distance} in \cite{28}. Results on this dataset are used for comparing the interpolation smoothness. 

\subsubsection{Digit Datasets} MNIST and USPS\cite{15} are two handwriting digit datasets. They both contain ten numbers but with entirely different styles. We regard the classes of the number as conditions and interpolating two different datasets in the latent space. They are used to compare the capability of models to interpolate two different domains with conditions.

\subsubsection{Argoverse \cite{56}} Argoverse Motion Forecasting dataset contains the driving trajectories of two vehicles as well as the road map information. We extracted the trajectory data of two interactive vehicles and then implemented several data augmentation methods on it. To check the effectiveness of these augmented trajectory dataset generated, three trajectory prediction algorithms are tested on both the augmented and original dataset. Moreover, six kinds of risky scenarios, which are likely to happen in a real traffic environment, are artificially generated based on the Argoverse dataset.

\subsubsection{Baselines} Two methods are selected as baselines for comparison. The first one is a vanilla VAE structure named MTG in \cite{62}, which has no AdaIN module and no fusion loss term. We term the second baseline as \textit{Perturbed} which is a trajectory augmentation method proposed in \cite{61}. This method fixes the start and end points and randomly disturbs the midpoint pose, and then fit a smooth trajectory to the perturbed point, the start and end points.

\begin{figure}[t]
\centering
\includegraphics[width=8.5cm]{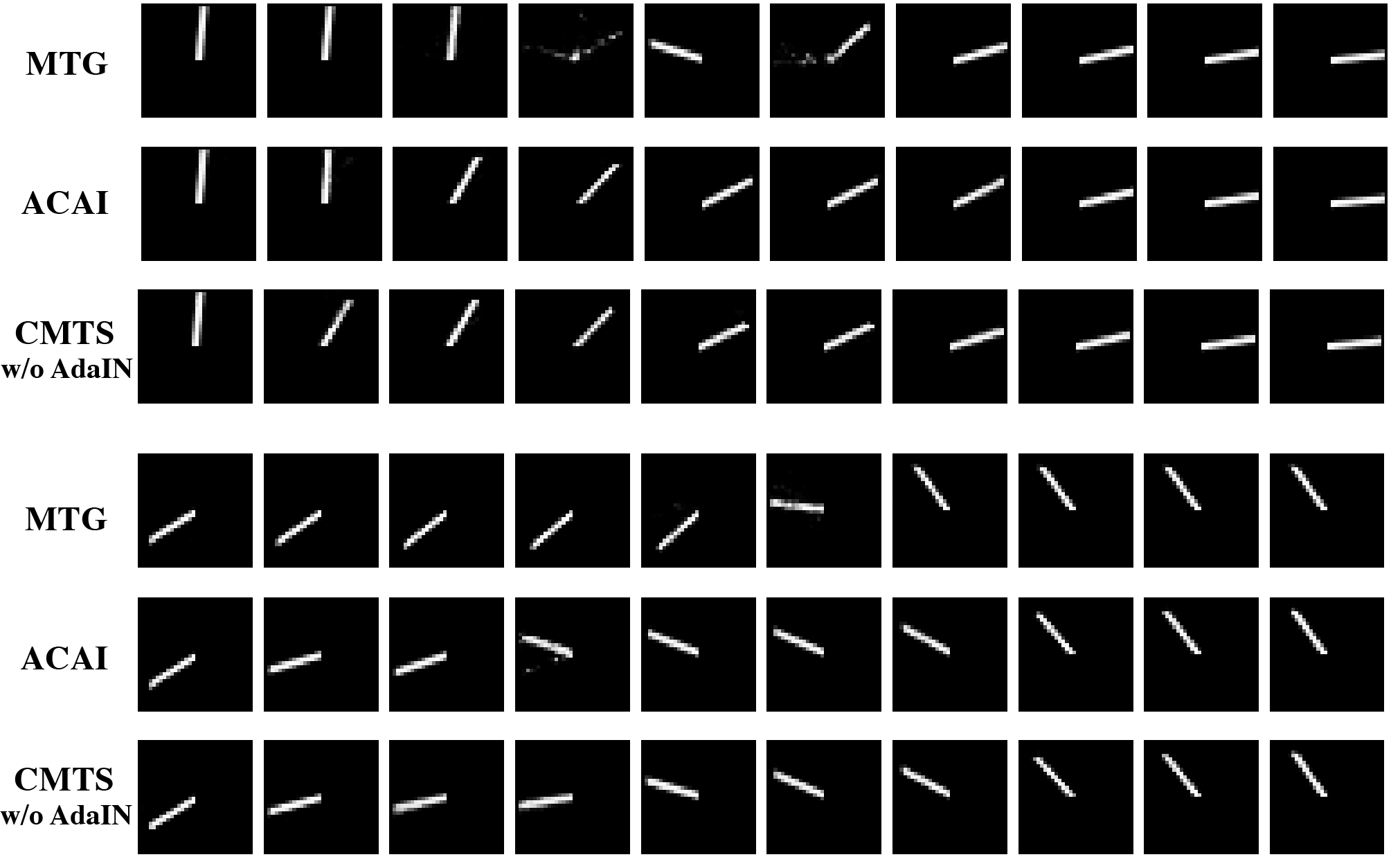}
\caption{Two interpolation examples of MTG, ACAI and CMTS on lines dataset. Images on the left and right are sampled from the dataset while the middle ones are generated. Since no label is provided, we only tested CMTS without style transfer AdaIN module.}
\label{lines}
\end{figure}

\subsection{Metrics}

\subsubsection{Cluster Number of Datasets}

To describe the complexity of a dataset, we use cluster numbers of the dataset as a metric for evaluation. Although information entropy is a common way to describe the complexity of one dataset, it requires labels that are unavailable in our case. Therefore, we leverage a non-parametric Bayesian method, Dirichlet Process Gaussian Mixture Models (DPGMM) \cite{57}, to cluster the dataset without a predefined cluster number $K$. The larger $K$ is, the more complex the dataset is.

\begin{table}[th]
\begin{center}
\caption{Metrics of Lines Dataset}
\label{lines_table}
		\begin{tabular}{c|c|c}
		\hline
		Metric          & Mean Distance ($\times 10^{-3}$) & Smoothness \\
		\hline
		VAE\cite{6}       & 1.21$\pm$0.17  & 0.49$\pm$0.13\\
		AAE\cite{58}      & 3.26$\pm$0.19  & 0.14$\pm$0.02\\
		VQ-VAE\cite{59}   & 5.41$\pm$0.49  & 0.77$\pm$0.02\\
		ACAI\cite{28}     & \textbf{0.24$\pm$0.01} & \textbf{0.10$\pm$0.01} \\
		\hline
		MTG          & 1.07$\pm$0.11  & 0.45$\pm$0.06\\
		CMTS w/o AdaIN & 0.32$\pm$0.02 & 0.15$\pm$0.01\\
		\hline
		\end{tabular}
\end{center}
\end{table}

\begin{table}[th]
\begin{center}
\caption{Dataset Complexity $^1$ of Argoverse Dataset}
\label{cluster}
		\begin{tabular}{c||c|c|c|c|c}
		\hline
		Dataset  & OD        & CD         & MTG       & CMTS       & OD+CMTS  \\
		\hline
		K        & 64$\pm$7  & 76$\pm$16  & 53$\pm$8  & 71$\pm$11  & \textbf{127$\pm$20}\\
		\hline
		\end{tabular}
\end{center}
\footnotesize{$^1$ Higher is better. All datasets have the same number of samples.}\\
\end{table}

\begin{figure}[t]
\centering
\includegraphics[width=8.5cm]{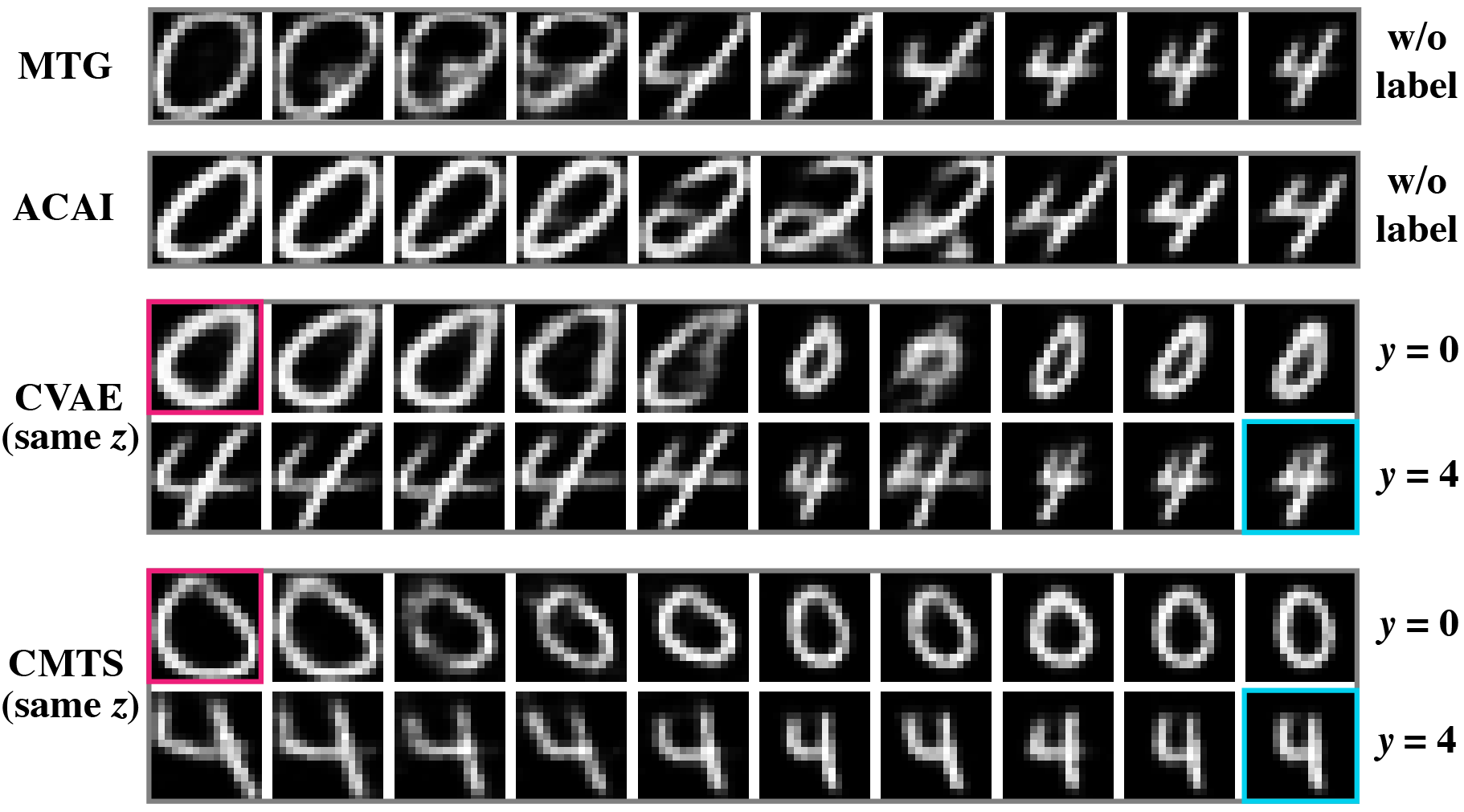}
\caption{Interpolation examples of digit datasets. For MTG and ACAI, no label information is used and the images on the left and right are sampled from datasets. For CVAE and CMTS, each row is conditioned on one style and the two images on upper left and lower right are sampled from datasets.}
\label{mnist}
\end{figure}

\begin{table*}[th]
\begin{center}
\caption{Error of Trajectory Prediction (2 seconds as history and 3 seconds of prediction)}
\label{mse}
		\begin{tabular}{c||c|c|c|c||c|c|c|c||c|c|c|c}
		\hline
		Method  & \multicolumn{4}{c||}{Vanilla-LSTM}&\multicolumn{4}{c||}{Social-LSTM \cite{16}} &\multicolumn{4}{c}{CS-LSTM \cite{20}}\\
		\hline 
		Dataset & OD & Perturbed & MTG   & CMTS    & OD & Perturbed & MTG      & CMTS           & OD & Perturbed & MTG   & CMTS \\
		\hline
		MSE     & 0.093    & 0.090     & 0.087 & \textbf{0.083}  & 0.074    & 0.072     & 0.067    & \textbf{0.062} & 0.048    & 0.046     & 0.046 & \textbf{0.042}\\
		MDV     & 0.025    & 0.035     & 0.081 & \textbf{0.143}  & 0.027    & 0.038     & 0.114    & \textbf{0.125} & 0.032    & 0.043     & 0.097 & \textbf{0.163}\\
		MDN     & 2.682    & 2.153     & 1.552 & \textbf{1.181}  & 2.732    & 2.321     & 1.463    & \textbf{1.029} & 2.124    & 1.852     & 1.274 & \textbf{0.961}\\
		\hline
		\end{tabular}
\end{center}
\end{table*}

\subsubsection{Trajectory Risk Metric}

The performance of trajectory prediction algorithms on the original dataset can be easily evaluated with mean square error (MSE), while on artificially designed near-miss trajectories, MSE is unable to capture the risk information. We propose to use the minimal distance between two vehicles (MDV) to represent the encounter risk and the mean distance between neighbor waypoints (MDN) to represent the smoothness.

%%%%%%%%%%%%%%%%%%
\section{Result Analysis}
%%%%%%%%%%%%%%%%%%

\subsection{Lines Dataset: Interpolation Performance}

The interpolation examples are displayed in Fig.~\ref{lines}. CMTS shows smoother results than VAE baseline and has similar results with the state-of-the-art interpolation method ACAI \cite{28}. In Table.~\ref{lines_table}, quantitative analysis is provided with two metrics proposed by \cite{28}. CMTS achieves a similar score with ACAI and outperforms other methods, which prove that CMTS is able to achieve smooth interpolation in simple situation with one attribute.

\subsection{Digit Datasets: Conditional Style Transfer}

Fig.~\ref{mnist} displays the results of interpolation on MNIST and USPS datasets using four methods. Since MTG and ACAI cannot leverage label information, we add Conditional VAE (CVAE) \cite{47} as a new competitor, which uses labels to control generated results. The result of MTG has a noisy boundary between two kinds of class since the style and content information is entangled. This entanglement makes it difficult to achieve a smooth interpolation. In contrast, CVAE achieves better results than MTG and ACAI because it separated the number class as a condition. However, our CMTS outputs better interpolation results than CVAE. The changing between two entire different styles is much smoother, which demonstrates that CMTS has the ability to fuse information from two domains with conditions.

\subsection{Argoverse: Increment of Data Complexity}

We evaluated four datasets with the DPGMM tool: the original Argoverse dataset (OD); a collision dataset (CD) obtained by translating two trajectories to a predefined collision point in OD; a dataset generated by MTG and a dataset generated by CMTS. In the generating stage of MTG and CMTS, the parameter $\lambda$ is fixed to 0.3 to synthesize samples that are closer to risk scenarios.

The datasets are first encoded into a latent space and then four non-parametric Bayesian models are trained using these low-dimensional codes from four datasets, respectively. We set the concentrating parameter $\alpha$ to 1 for all models and use the output cluster number $K$ as an indicator of the complexity of the datasets. The results are shown in Table.~\ref{cluster}.
It shows that the dataset generated by CMTS contains larger numbers of clusters than OD and CD. The combination of CMTS dataset and OD achieves the highest number of clusters among all datasets, which means CMTS dataset contains clusters unseen in OD and CD, otherwise CMTS+OD will have similar $K$ as OD.

\begin{figure}[t]
\centering
\includegraphics[width=8.5cm]{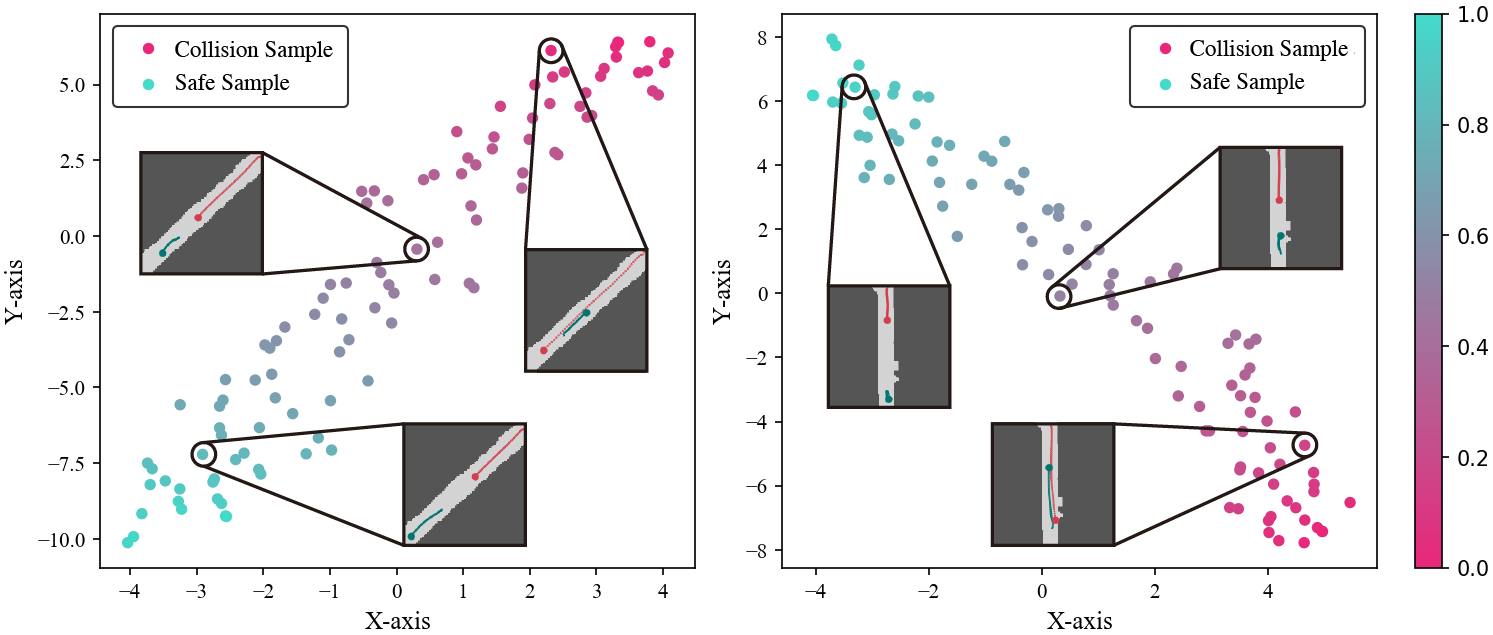}
\caption{T-SNE \cite{60} results of interpolation samples. Points with pink color and cyan color represent the safe samples and the collision samples, respectively.}
\label{tsne}
\end{figure}

To visualize the rare trajectories generated by CMTS, Fig.~\ref{transfer} gives three examples, each of which has two map conditions from the original and collision datasets. Besides, we use T-SNE \cite{60} to display the linear interpolation samples in a 2-dimensional space in Fig.~\ref{tsne}, where the interpolation samples roughly line up in the 2-dimensional space.

\subsection{Argoverse: Improvement of Trajectory Prediction}

Three trajectory prediction methods (Vanilla-LSTM Social-LSTM \cite{16} and CS-LSTM \cite{20}) and two kinds of driving scenarios are selected to test the augmented dataset. 
The first kind is safe driving scenarios sampled from the original Argoverse dataset. The second kind is risky situations that are rarely collected in the naturalistic dataset. Therefore, we artificially design six risky scenarios based on Argoverse dataset (Fig.~\ref{danger}). Details about the design process can be found in the supplementary material. 

In Table.~\ref{mse}, we compare the performance of prediction algorithms trained on four datasets: original dataset (OD), Perturbed dataset (augmented by random perturbation), datasets generated by MTG and CMTS. For all three trajectory prediction models, CMTS achieves the best result in MSE, MDV and MDN, which proves that new scenarios help trajectory prediction algorithms to improve their capability of dealing with risky situations.

To evaluate the performance in risky scenarios, we selected 18 examples predicted from CS-LSTM as shown in Fig.~\ref{danger}. Since the near-miss data does not exist in OD, the predictions from the model trained on OD are awful. The reason is that the risky scenario data comes from other domains, which has exceeded the generalization scope of the model trained only on OD. For the two models trained on MTG and CMTS, we observe that the model trained on CMTS dataset predicts more smooth and reasonable trajectories.

\begin{figure}[t]
\centering
\includegraphics[width=8.5cm]{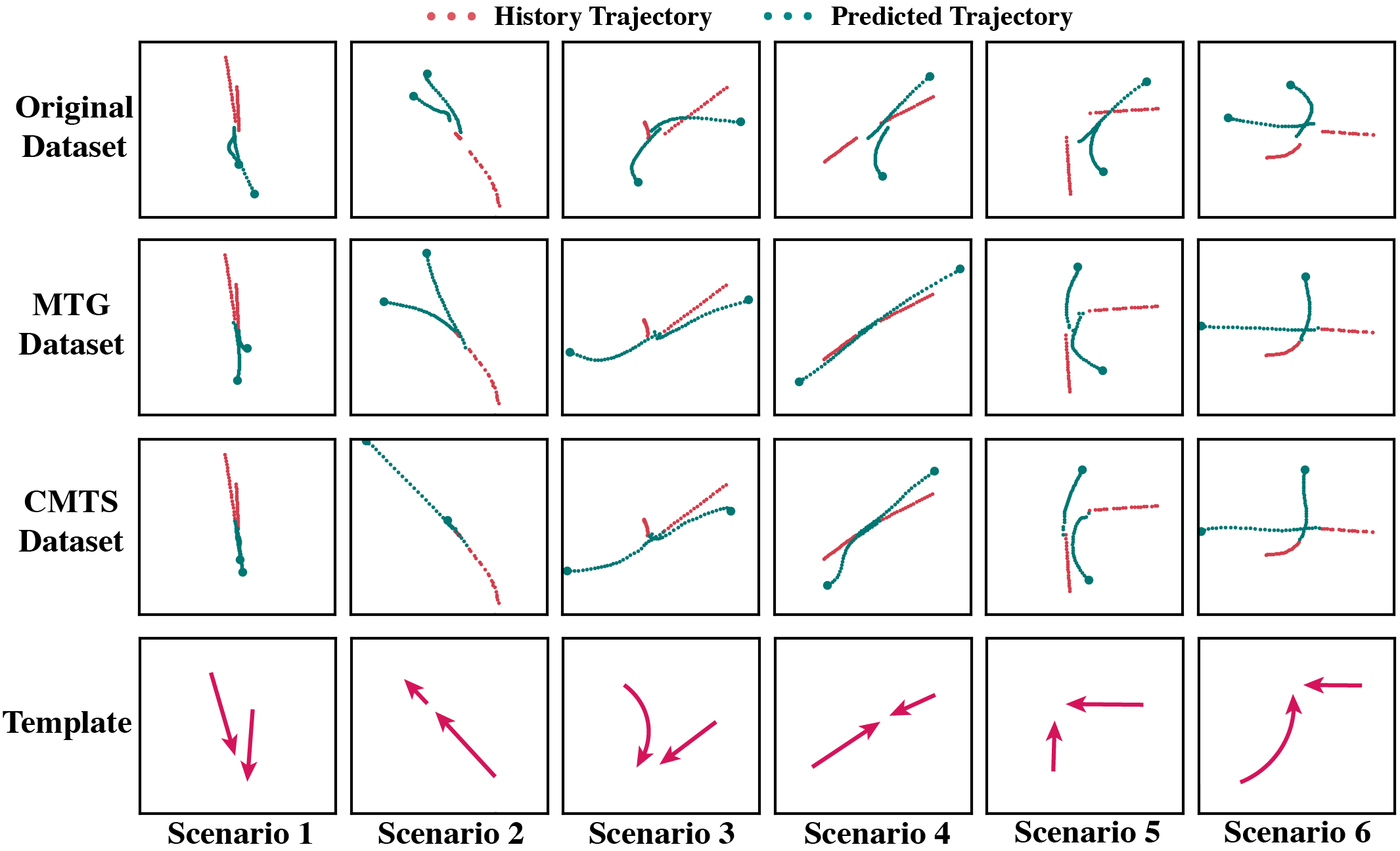}
\caption{Trajectory prediction results in six risky scenarios with CS-LSTM \cite{16}. Red points and green points represent the history and predicted trajectory, respectively. The large green points are the ends of predictions.}
\label{danger}
\end{figure}

%%%%%%%%%%%%%
\section{Conclusion}
%%%%%%%%%%%%%

To improve the accuracy of trajectory prediction algorithms and their capability of dealing with risky driving scenarios, a data augmentation method, CMTS, based on Variational Bayesian method, is proposed in this paper to generate rare and near-miss traffic trajectories. Experiments show that the augmented data increases the diversity and complexity, and constructs rare near-miss scenarios that do not exist in the original dataset. Moreover, these synthetic data can improve the accuracy of state-of-the-art trajectory prediction algorithms. 

We believe synthesizing rare and risky data is valuable because firstly, such data in the real world can hardly be collected, and secondly, even in the simulator, it is very complex and difficult to construct such a rare scenario. The generated near-miss data can not only help improve the accuracy of forecasting algorithms but also can be used to evaluate various types of path planning and control algorithms, so as to improve the ability of the entire pipeline of autonomous driving algorithm to deal with risky driving scenarios and further improve safety.

In this work, we did not apply any physical and kinematic constraints to the generated trajectory, which is also important to make it closer to real trajectory. We plan to embed these constraints in our future work.

% We are able to control the feasible area of the generated trajectory according to the condition of the road constraint

\section*{Acknowledgment}
This research was sponsored in part by Uber ATG. The authors would like to thank Uber ATG team for valuable discussion.

\appendix
Hyper-parameters, network architectures and detailed experiment settings are described in the supplementary material: \url{https://wenhao.pub/publication/rare-supplement.pdf}.

%%%%%%%%%%%%%%
\bibliographystyle{IEEEtran}
\bibliography{main.bib}
%%%%%%%%%%%%%%

\end{document}